\documentclass[sigconf]{acmart} % Removed 'anonymous,review' for displaying authors

\usepackage{algorithm}
\usepackage{algorithmic}
\usepackage{pgfplots,subcaption}

\copyrightyear{2025}
\acmYear{2025}
\setcopyright{rightsretained}
\acmConference[KDD '25]{Proceedings of the 31st ACM SIGKDD Conference on Knowledge Discovery and Data Mining V.1}{August 3--7, 2025}{Toronto, ON, Canada}
\acmBooktitle{Proceedings of the 31st ACM SIGKDD Conference on Knowledge Discovery and Data Mining V.1 (KDD '25), August 3--7, 2025, Toronto, ON, Canada}
\acmDOI{10.1145/3690624.3709332}
\acmISBN{979-8-4007-1245-6/25/08}

\makeatletter
\gdef\@copyrightpermission{
  \begin{minipage}{0.2\columnwidth}
   \href{https://creativecommons.org/licenses/by/4.0/}{\includegraphics[width=0.90\textwidth]{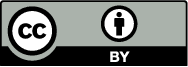}}
  \end{minipage}\hfill
  \begin{minipage}{0.8\columnwidth}
   \href{https://creativecommons.org/licenses/by/4.0/}{This work is licensed under a Creative Commons Attribution International 4.0 License.}
  \end{minipage}
  \vspace{5pt}
}
\makeatother

\begin{document}

\date{}

\title{Adaptive Domain Inference Attack with Concept Hierarchy}

\author{Yuechun Gu, Jiajie He, Keke Chen}

\email{{ygu2,jiajieh1,kekechen}@umbc.edu}

\affiliation{%
 \department{Trustworthy and Intelligent Computing Lab (TAIC), Computer Science and Electrical Engineering}
    \institution{University of Maryland, Baltimore County}
    \city{Baltimore}
    \state{MD}
    \country{USA}
}

\begin{abstract}
With increasingly deployed deep neural networks in sensitive application domains, such as healthcare and security, it's essential to understand what kind of sensitive information can be inferred from these models. Most known model-targeted attacks assume attackers have learned the application domain or training data distribution to ensure successful attacks. Can removing the domain information from model APIs protect models from these attacks? This paper studies this critical problem. Unfortunately, even with minimal knowledge, i.e., accessing the model as an unnamed function without leaking the meaning of input and output, the proposed adaptive domain inference attack (ADI) can still successfully estimate relevant subsets of training data. We show that the extracted relevant data can significantly improve, for instance, the performance of model-inversion attacks. Specifically, the ADI method utilizes the concept hierarchy extracted from the public and private datasets that the attacker can access and applies a novel algorithm to adaptively tune the likelihood of leaf concepts in the hierarchy showing up in the unseen training data. For comparison, we also designed a straightforward hypothesis-testing-based attack -- LDI. The ADI attack not only extracts partial training data at the concept level but also converges fastest and requires the fewest target-model accesses among all candidate methods. Our code is available at \url{https://anonymous.4open.science/r/KDD-362D}. 
\end{abstract}


\maketitle

\section{Introduction}
\label{introduction}
Large-scale deep learning models are increasingly deployed in application domains, playing pivotal roles in sectors where sensitive or proprietary data is used in training the models \cite{romero2021, shameer2017, Alspector_Dietterich_2020}. These models might be packaged in API services or embedded into applications, becoming a concerning new attack vector. Recent studies have shown web services are exposed to various types of attacks \cite{Diaz21, Ibrahim20}. One such practical attack is stolen secret credentials or API secret ID. With the stolen credentials, several effective model-targeted privacy attacks, including model inversion (or training data reconstruction) \cite{fredrikson2015,zhang2020}, property inference \cite{ ganju2018,zhang2021}, and membership inference attacks \cite{shokri17,hu2022}. 

However, we have noticed all these model-targeted attacks depend on a certain level of knowledge about the application domain. (1) Model-inversion (MI) attacks depend on a learning procedure, e.g., a GAN method \cite{zhang20}, to progressively adjust seed images from a domain or distribution similar to the training data domain towards most likely training examples. Without this domain knowledge, i.e., with irrelevant auxiliary data, the attack performance can be significantly reduced \cite{gu2022}. (2) Membership-inference attacks (MIA) estimate the possibility of a target sample belonging to the training data of a model. Most MIA attacks\footnote{Most recently, Carlini et al. \cite{carlini2022lira} propose to use one-side non-member hypothesis testing for approximately identify the membership, which may not need to know the target distribution. However, knowing the domain will help attackers quickly identify true positive samples.} assume attackers know the type and distribution of training to train shadow models, which are also important for making sense the MIA result in practice. (3) Property-inference attacks try to uncover global properties of the training data that the model's creator did not intend to reveal. These attacks also need shadow classifiers that are trained on tasks similar to that of the target classifier. Domain knowledge plays an important role in training shadow classifiers.

Thus, one may wonder whether it can protect a model from all these attacks by stripping off the domain-related information in model applications. More specifically, can we achieve the protection goal by packaging the model as an unnamed function call or a service API: $y = f(x)$, where an input $x$ leads to a prediction $y$, and no meaning of $x$ and $y$ is given? Such a service API mechanism is still convenient and deployable in a cloud-based application ecosystem. 

Our recent study shows that even with such a minimal-knowledge setting, attackers can find ways to re-establish knowledge about the application domain. We \cite{gu2022} first attempted to reconstruct the domain information with model accesses only. We used a generative adversarial network (GAN) approach (GDI) to infer which of a set of known candidate datasets is most similar to (or likely from) the target model's domain. 

However, the GDI method has several drawbacks. (1) It's dataset-oriented, i.e., assuming datasets similar to the target training data exist in the candidate datasets. Our experiment shows that this method does not work well if the candidate dataset is only partially similar to the target dataset, e.g., only one or a few classes are similar. (2) The GAN-based procedure requires excessive accesses to the target model, which may alarm the model service owner. 

\textbf{Scope of our research.} While GDI targets the dataset-level inference, we wonder whether the record-level inference, i.e., membership inference without knowing the domain, is also possible. We noticed that the offline version of the recently developed Likelihood Ratio Attack (LiRA) \cite{carlini2022lira} meets our requirements, which only needs to train offline shadow models without domain knowledge. We name this first candidate method: offline LiRA domain inference (LDI). However, we found that simply applying LiRA does not work satisfactorily for two reasons. First, it assumes that all out-domain samples perform similarly following a normal distribution regardless of target domain distributions, which may lead to errors for specific target domains. Second, it needs to test every provided record to determine its likely membership, which is expensive. We experimented with improved methods like class-wise sampling, but the result was still unsatisfactory.   

By extending the idea of record-level inference to class-level inference, we propose the \emph{adaptive domain inference attack} (ADI) to address the weaknesses of both GDI and LDI. Specifically, the ADI approach can detect likely training examples at the concept (or class) level. ADI makes the extracted domain information more accurate than the GDI's dataset level. ADI also utilizes the concept hierarchy to direct the inference to stay focused on likely concepts, which avoids inefficient record-level procedures like LDI. We find that ADI requires much fewer model accesses than GDI and LDI and yields better-quality inference, as Section \ref{sec:access_exp} shows. 

The ADI attack assumes that the attacker has minimal knowledge about the unnamed model API, e.g., a black-box image classification model with only information about the input image size and output probability vector. The attacker tries to estimate the target domain via the existing knowledge about a (possibly relevant) mini-world in the form of concept hierarchy.  The process can be described as follows. First, the ADI attack establishes a concept hierarchy from the data pool that the attacker can collect or adopts an existing one, e.g., the ImageNet concept hierarchy (Figure \ref{fig:concept2} in Appendix C), as long as samples at the leaf concepts are available. Each leaf node of the concept hierarchy represents a concept associated with a cluster of sample images. Each internal node represents a higher-level concept, e.g., a group of relevant concepts. Each node is also associated with the probability of being relevant to the target model's domain, initially set to equal probability. The ADI algorithm will iteratively adjust the node probabilities with the samples from the randomly sampled leaf concepts and the feedback from the target model prediction. After a few iterations, the probabilities of the relevant leaf concepts are promoted, while irrelevant ones get demoted. Finally, we sample leaf concepts according to their probabilities to generate a candidate dataset to assist model-based attacks.

Our method has several unique advantages. (1) It can identify relevant classes of records among a huge data pool more accurately and efficiently than the dataset-oriented GAN-based domain inference attacks (GDI) \cite{gu2022} and our record-oriented LDI method. The experimental results also show that the hierarchical concept organization is crucial in guiding the algorithm to identify the relevant concept more effectively than flatly organized concepts. In assisting model-inversion attacks, the ADI-extracted datasets work about 20\% better than the candidate datasets identified by GDI. (2) It achieves better attacking results with much fewer model accesses. It works on a small batch of samples per iteration, e.g., 1000 images, and the number of iterations is small, e.g., around 40. Overall, ADI requires orders of magnitude fewer model accesses than GDI and LDI.

The remaining sections include the basic notations and definitions (Section \ref{preliminery}), the detailed description of the LDI and ADI attacks (Section \ref{method}), the experimental evaluation (Section \ref{sec:experiments}), the related work (Section \ref{sec:related work}), and our conclusion (Section \ref{sec:conclusion}). 

\section{Prelimineries}
\label{preliminery}

%This section introduces the definitions and necessary background knowledge about domain similarity measures.

\subsection{Definitions} 

In our context, the machine learning model under attack, denoted as $f(x)$, processes input data $x$ (such as an image) and outputs a confidence vector, $v=f(x)$. This vector contains the probabilities that the input belongs to each potential class. The label, $y$, corresponds to the class with the highest probability in this vector. The model is trained using a dataset, $D_T$, a subset drawn from an unknown ``Latent Domain’’ $S_T$. Once the model is trained and deployed, users interact with it, typically via an API. They don't have any direct access or knowledge about the original training data, $D_T$. 

\textbf{Data pool.} A data pool may consist of multiple training datasets from public or private domains, $P = \{ D_1, D_2, \dots, D_n\}$, where $D_i$ contains feature vector and label pairs. The attacker prepares a data pool to perform the attack. 

\textbf{Dataset similarity.} Our proposed attack extracts data records, $D_e$, from the data pool that are closely related to the target training data $D_T$. The effectiveness of the attack is measured by the similarity of $D_e$ to $D_T$, i.e., Distance($D_e, D_T$). This similarity is crucial for effective model-based attacks, such as model-inversion attacks \cite{zhang2020,gu2022}. We use the Optimal Transport Dataset Distance (OTDD) \cite{alvarez2020geometric} to assess this similarity. OTDD measures dissimilarity between datasets using optimal transport distances, providing geometric insight and interpretable correlations.

\section{Adaptive Domain Inference Attack}
\label{method}
The proposed attack aims to directly identify the classes of records in a data pool similar to the training data of the target model. The extracted samples can be used as auxiliary data in model-based attacks. We first discuss the threat model, then briefly describe the record-level LiRA domain inference attack (LDI), and finally present the motivation and detail of ADI. 

\subsection{Threat Modeling}
\label{sec:threat modeling}
\textbf{Involved parties.} The model owner may package the deep learning model as a web service and remove all semantic information from the input and output. The adversary can be any party who is curious about the model and the data used for training the model.

\textbf{Adversarial knowledge.}  Web service APIs are secured by standard protocols, i.e., via HTTPs (HTTP over TSL or SSL), which ensures communication security. The attacker does not breach the protocol and encryption but uses the stolen credentials to issue regular API calls. The model API accepts only encrypted inputs and outputs, not containing any domain information. We assume attackers have successfully stolen the credentials and have black-box access to the model with limited knowledge about API input/output information, such as input image size and the number of output classes. Our attack focuses on image classification models, with the API returning a class likelihood confidence vector. However, the attack can be extended to non-image tasks. For label-only outputs, attackers can use strategies like \cite{Choquette-Choo21} to create pseudo-confidence vectors. Attackers can collect images from diverse sources to build the concept hierarchy, enhancing attack coverage. Larger and more diverse sources increase the likelihood of containing concepts used by the target model, but direct access to the target-domain training and testing datasets is not possible.

\textbf{Attack target.} Domain information, such as image types and class definitions, is crucial for model-based attacks. Knowing the domain helps adversaries select suitable auxiliary datasets, enhancing attack efficacy \cite{zhang2020,matsumoto2023,hayes2017}. Our goal is to determine if this information can be derived solely from the exposed model API. Attack performance (relevance of generated data to the original data) and attack cost (number of accesses to the target model) are the main measures for domain inference attacks. The first such attack, GDI \cite{gu2022}, is expensive and performs poorly if landmark datasets are only partially relevant to the target domain.

\subsection{First Candidate -- LDI}
\label{sec:LiRA}
We wonder whether we can conduct a direct record-level inference without the domain knowledge. Specifically, can we use a membership inference attack (without knowing the domain!) to collect the likely in-domain records and then analyze them to infer the domain information? Most MIA attacks assume the known domain and depend on it to derive shadow models \cite{hu2022}, except for a most recent development: the offline version of likelihood ratio attack (LiRA) \cite{carlini2022lira}. Next, we explore the application of offline LiRA in domain inference and summarize its problems.

\textbf{Offline LiRA test.} Carlini et al. \cite{carlini2022lira} present both an online and an offline LiRA test. The online LiRA test requires attackers to train multiple shadow models on both in-domain and out-domain datasets, resulting in extremely high computational costs. In contrast, the offline LiRA test does not use in-domain information and relies solely on out-domain samples. More details can be found in the paper \cite{carlini2022lira}. This offline version is well-suited for our domain inference attack, as it does not assume any target domain is known. 

Specifically, to test the membership of $x$, we measure the probability of observing confidence as high as the target model's under the null hypothesis that the target point \((x, f(x))\) is a non-member as follows:

\begin{equation}
\label{eq:estimate}
\Lambda(x)  = 1 - \Pr[Z > \phi(q)], \text{ where } Z \sim \mathcal{N}(\mu_{\text{out}}, \sigma_{\text{out}}^2)
\end{equation}

where $q$ represents the difference between the greatest and second greatest confidence scores of the predicted logits. The function $\phi(q) = \log\left(\frac{1}{1-q}\right)$ applies logit scaling, and $\mathcal{N}()$ denotes a Gaussian distribution.
A one-sided hypothesis testing is conducted to conclude whether the confidence is high enough to reject the null hypothesis, as a member sample’s $\phi(q)$ value is higher and significantly out of the out-domain samples’ $\phi(q)$ distribution. For simplicity, $\Lambda > 0.5$ has been used as the threshold to determine the membership of in-domain samples \cite{carlini2022lira}.

\textbf{Offline LiRA domain Inference.} We can apply the offline LiRA to a record-based domain inference attack as follows. We start with training $n$ shadow models on $n$ hypothetical in-domain shadow datasets, each constructed by randomly picking samples from the data pool by flipping a coin. For each shadow dataset, we have the remaining samples in the data pool as out-domain samples, which are used to estimate the parameters $\mu_{out}$ and $\sigma_{out}^2$.

Then, we test the membership of each instance in the data pool for the target domain. The instance is used as the input to the target model to generate the corresponding $\phi(q)$, and we apply the offline LiRA to determine the membership. Once (likely) member instances are collected, the attacker can analyze them to derive the domain information, e.g., based on the top-K most popular classes or clusters.

We also tested two sampling methods to minimize the model access cost: random sampling and class-balanced sampling. In random sampling, we uniformly randomly select a subset of data records from the data pool. In class-balanced sampling, we take the same number of samples from each class to avoid oversampling large classes and increase the coverage of small classes. It's specifically implemented by augmenting the less populated classes, i.e., via random rotation and flipping of existing samples in the class. The effect of different sampling strategies is shown in Section \ref{sec:LDI_result}. Detailed steps of LDI are provided in Algorithm \ref{alg:LDI}.

\begin{algorithm}[H]
\caption{LiRA-based domain inference attack}
\label{alg:LDI}
\begin{algorithmic}[1]
\REQUIRE Model $f_{D_T}$, data pool $D_A$
\STATE $\text{confs}_{\text{out}} \gets \{\}$ 
\STATE $\tilde{D}_T \gets \{\}$ 
\FOR{$i = 1$ to $N$}
    \STATE $D_{\text{shadow}} \leftarrow D_A$ \COMMENT{Sample a shadow dataset}
    \STATE $f_{out} \leftarrow \mathcal{T}(D_{\text{shadow}} \setminus \{(x, y)\})$ \COMMENT{Train OUT model}
    \STATE $q \leftarrow f_{out}(x)_1 - f_{out}(x)_2$ \COMMENT{Get an observation}
    \STATE $\text{confs}_{\text{out}} \leftarrow \text{confs}_{\text{out}} \cup \{\phi(q)\}$
\ENDFOR
\STATE $\mu_{\text{out}} \leftarrow \text{mean}(\text{confs}_{\text{out}})$
\STATE $\sigma_{\text{out}}^2 \leftarrow \text{var}(\text{confs}_{\text{out}})$

\FORALL{$(x,y) \in D_A$}
    \STATE $q \leftarrow f_{D_T}(x)_1 - f_{D_T}(x)_2$ \COMMENT{Get an observation}
    \STATE $\Lambda \gets 1 - \Pr[Z > \phi(q)]$, where $Z \sim \mathcal{N}(\mu_{\text{out}}, \sigma_{\text{out}}^2)$
    \IF{$\Lambda \geq 0.5$}
        \STATE $\tilde{D}_T \leftarrow \tilde{D}_T \cup \{(x,y)\}$
    \ENDIF
\ENDFOR
\end{algorithmic}
\end{algorithm}

Despite using this refined design, we observed that LDI performs unsatisfactorily due to several reasons. First, the offline LiRA method assumes all out-domain instances' output confidence values $\phi(q)$ fit in the normal distributions, which may not be precise, as discussed in the original paper \cite{carlini2022lira}. Second, if the target domain instances form only a small portion of the data pool, neither sampling method may ideally extract the likely target domain instances. We show more details in Appendix A. 

\subsection{Concept Hierarchy for ADI} 
\label{sec:concept_Hierarchy_total}
\begin{figure}[h] % figure placement: here, top, bottom, or page
   \centering
       \includegraphics[width=\linewidth]{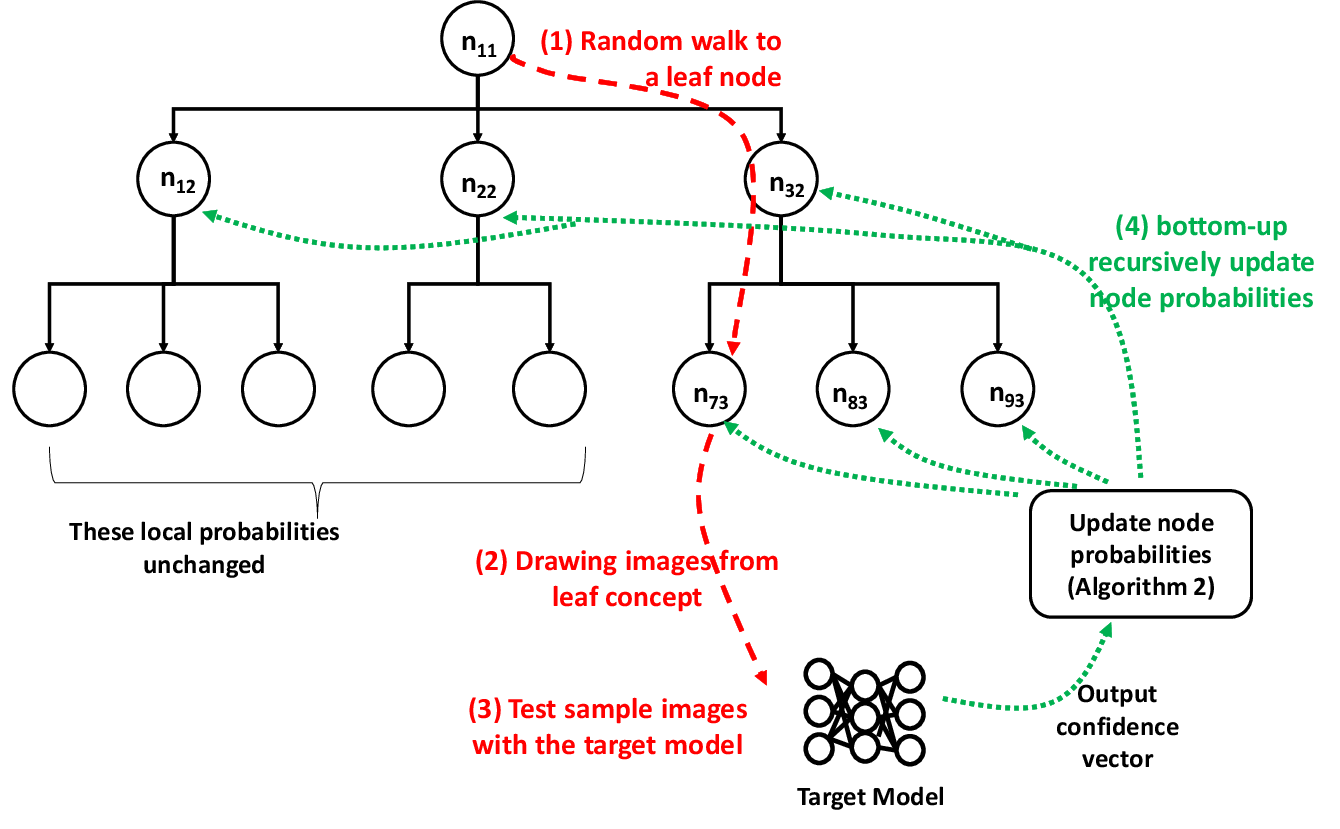} 
       \caption{Concept hierarchy illustration. The $i$-th node at level $j$ holds probability $p_{ij}^{(t)}$ at time step $t$. Initially, node probabilities are set to $1/q$, with $q$ as the number of child nodes under the node's parent. In each iteration, Algorithm \ref{alg:general} applies a random walk from the root to a leaf, e.g., a red path. Then, a sample image is drawn from the leaf node and applied to the target model. The result will trigger the probability updates following a green path from the leaf to the root.}
       \label{fig:tree-like hierarchy}
\end{figure}

Bearing the problems with LDI in mind, we introduce the adaptive domain inference based on concept hierarchy (ADI).
The use of concept hierarchy, depicted in Figure \ref{fig:tree-like hierarchy}, is pivotal to our approach, steering the attack towards probable classes. We will show that a concept hierarchy can help the algorithm better focus on the relevant concepts (and their branches), which is much more effective than a flat concept organization used by LDI. It also serves application scenarios better, as an application dataset likely uses multiple relevant concepts, which can be better captured by the hierarchical structure. 

We assume a concept hierarchy has been derived from the data pool or borrowed from other applications, while the hierarchy construction method is not the focus of this research. The concept hierarchy is expected to be large enough so that it can intersect or even contain the concepts used by the target model's potential domain. This hierarchical structure spans $L$ levels with the root at level 1, and each branch groups similar samples, where naming might not be important. For example, an ``skirt'' node under ``clothes'' may lead to subclasses ``hoop skirt'' and ``mini skirt'' nodes, while ``skirt'' under ``beef'' represents the images of a specific part of beef. Each node carries a \emph{local probability}, starting at $1/q$ for $q$ siblings. Traversing from the root to a leaf and multiplying the probabilities on the path, we can get a leaf's \emph{global probability}. These leaf global probabilities sum to 1, maintaining the algorithm's integrity. As the attack progresses, the algorithm adjusts local probabilities based on the target model's feedback, culminating in leaf nodes' global probabilities that reflect their presence in the target model's training data.

We have experimented with two simple ways to obtain the concept hierarchy. First, attackers can use an existing image-based concept hierarchy, e.g., one built from the ImageNet \cite{deng2009} repository, which we have used to show the attack on the DeepFashion model in our experiments. Second, attackers may have an initial guess and collect some datasets that are likely relevant to the target domain. A straightforward approach is to leverage the inherent structure of the collected data pool $\{D_1, D_2, \dots, D_n\}$ to build a 3-level tree. The nodes at the second level represent individual datasets, and their child nodes correspond to the associated classes or clusters within those datasets. With these concept hierarchies, we have observed ADI performs significantly better than other candidate methods. We are sure that more refined concept hierarchies will further improve our algorithm’s performance.

\subsection{Details of ADI Attack }

\label{sec:ADI}

Once the attacker has assembled a data pool and formulated a concept hierarchy, they use Algorithm \ref{alg:general} to adjust the node-associated probabilities. The core of the algorithm is an iterative process, as illustrated in Figure \ref{fig:tree-like hierarchy}. Each iteration involves drawing sample images from leaf clusters, using a top-down random walk following the nodes' local probabilities. The sampling process starts from the root and randomly selects a child branch based on their local probabilities until a leaf is reached. Each sample is then fed into the target model $f(\dot)$ that produces a confidence vector, $v$.

The crux is how to use the confidence vector to tune the node probabilities. The first step is to determine whether the node is relevant enough to the domain, for which we design two methods. 

(1) \textbf{LiRA-based method} (LiRA-ADI): We use the offline LiRA test directly to determine the likelihood of membership for the test sample, $\Lambda$. If $\Lambda\geq0.5$, we label the sample as a \emph{positive sample}. However, the LiRA test requires a significant setup cost, i.e., training multiple shadow models. Thus, we design the following alternative method. 

(2) \textbf{Entropy-based method} (Entropy-ADI): The second method is built on the intuition that if an image is similar to a training data class, the confidence probability $v_i$ for that class, $y_i$, significantly exceeds others; if the model fails to recognize the image, class probabilities tend to be similar. We capture this trait by using the concept of \emph{normalized entropy}, and employing a predefined entropy threshold $\lambda$ to determine if the target model confidently recognizes the sample. Since the range of possible entropy values is determined by the number of classes, $m$, i.e., $[0, \log_2 m]$, the normalized entropy function is
\begin{equation}
\label{eq:Entropy}
\tilde{H}(v) = -\frac{1}{log_2 m}\sum_{i=1}^m v_i log_2 v_i,
\end{equation}
where $v = (v_1,\dots,v_m)$ is the confidence vector. The normalization converts all entropy values, regardless of the number of classes,  to the range $[0, 1]$, and allows us to establish a generalized algorithm, independent of the target model's class count. We label samples with entropy $\le \lambda$ as \emph{positive samples} since a target-model-recognized sample typically has a low-entropy confidence vector, and those with entropy $>\lambda$ as \emph{negative samples}. In experiments, we have observed $\lambda=0.83$ seems to give the best result. 

To unify these two methods, we define the $positive(x, mode)$ function with \emph{mode} indicating LiRA-ADI or Entropy-ADI, which tests whether a sample $x$ is positive. Consequently, we reward the leaf node from which a positive sample was drawn by increasing the leaf's local probability. The node probability adjustment is then propagated to sibling and parent nodes, as detailed in the following section. Conversely, a negative example results in a decreased probability for its originating leaf cluster, and the adjustment is also propagated to sibling and parent nodes. Through rounds of these adjustments, we hope that the leaf probabilities are stabilized, reflecting their relevance to the hidden domain of the target model.

\begin{algorithm}[H]
\caption{Overview of ADI ($f()$, $mode$, $C$, $b$)}
\label{alg:general}
\begin{algorithmic}[1]
\REQUIRE Target model $f()$, $mode$: LiRA-ADI or Entropy-ADI, Batch size $b$, 
         Concept hierarchy $C$ with local probabilities attached to the nodes.
\ENSURE Updated concept hierarchy $C$

\STATE $t \gets 0$
\REPEAT
    \STATE \textbf{Perform random walks on $C$ from the root to leaves $b$ times, 
           and draw $b$ images accordingly. Current batch of images: $I_t$}
    \FOR{$k = 1$ to $b$}
        \STATE $i \gets$ leaf node index from which $I_{t,k}$ is drawn
        \STATE Update $C$ using $\text{Adj\_Probs}(C, f(), mode, s, i)$
    \ENDFOR
\UNTIL{Convergence is achieved for $I_t$}

\RETURN $C$
\end{algorithmic}
\end{algorithm}

The next subsections will give more details for the core steps: the node probability adjustment strategy: Adj\_\,Probs $(C, f(), $mode$, I_{tk}, i)$, and the convergence condition: converge($I_t$).

\subsubsection{Concept-probability Adjustment and Propagation} 
\label{sec:concept adjustment}
The algorithm's central step uses the target model's output to modify node probabilities within the concept hierarchy. Positive feedback escalates the probability of the originating leaf cluster and its neighboring ones, while negative feedback reduces them. 

To clearly describe the probability adjustment and propagation procedure, we give the following notations first. We denote the $i$-th node (from left to right) at the $j$-th level ( $1..L$ from top to down) of the concept hierarchy as node $n_{ij}$. As such, each node can be uniquely identified.  Let the probability associated with the node at iteration $t$ be $p^{(t)}_{ij}$. The number of siblings of the node can vary -- for simplicity, we denote the siblings of the node as a set $U_{ij}$ and the number of siblings as $|U_{ij}|$. We also assume a sampled record, $s$, is drawn from the leaf node $n_{ iL}$ at timestep $t$. The following update will be recursively applied to the leaf node and its ancestors until the root is reached. 

\textbf{Target-node probability update.} Let $\delta (j)$ denote a level-wise adjustment. If the sample is positive, the adjustment is added to the probability of node $n_{ij}$; otherwise, the probability is decreased by $\delta(j)$. 
With the previously defined $positive (s, mode)$ function for sample $s$, we define: $w(s, mode) = positive(s, mode)? 1: -1$ and

\begin{align}
p^{(t)}_{ij} &= p^{(t-1)}_{ij} + w(s, mode)\delta(j) 
\end{align}

\textbf{Sibling probability update.} Correspondingly, we need to update the siblings' probabilities to maintain the sum of local probabilities under the parent to be 1. For each positively adjusted node, its siblings' probabilities will correspondingly decrease, and vice versa. 

\begin{align}
p^{(t)}_{ij} &= p^{(t-1)}_{ij} - w(s, mode) 
\frac{\delta(j)}{|U_{ij}|}, \quad \text{for $j \in U_{ij}$} 
\end{align}

Recursively, the target node is moved up to the parent of node $n_{ij}$, and the same target node and sibling probability update procedure is applied until the root is reached.  

\subsubsection{Adaptive adjustment $\delta$ and Probability Rebalancing} 
The bottom-up probability adjustment will progressively increase the probabilities of the concepts (and their parents) relevant to the target domain. The updated node probabilities will immediately affect the images sampled in the next iteration in Algorithm \ref{alg:general}.  

The amount of adaptive adjustment $\delta$ is designed to attain optimal convergence as detailed in Section \ref{sec:converge}. It also controls the intensity of probability update crossing levels. The setting of $\delta$ is a delicate balance among convergence quality, convergence speed, and the overall coverage of relevant concepts. If it's too large, the optimal convergence is difficult to reach, and the algorithm may be biased towards a few early visited concepts. If it's too small, convergence might be slow, increasing the attack's cost (i.e., the number of model API accesses). The heuristic we use, which has proven successful in experiments, ensures that $\delta(j)$ decays linearly towards the level $j$: the higher the level (the smaller $j$), the less the adjustment is propagated. Moreover, the more nodes the hierarchy has, denoted as $|C|$, the less dramatic the adjustment should be to increase the coverage of concepts. We have experimented with $\delta(j) = j/|C|$ with different scaling factors (Section \ref{sec:parameters}) and shown that $\delta(j) = j/|C|$ is appropriate.

However, merely using the above heuristic is not enough. We observed that aggregation over rounds of adjustment may severely unbalance node probability distributions. To address this, we discount the probability $p^{(t)}_{ij}$ surpasses the sum of its siblings' probabilities: $\sum_{k \in U_{ij}} p^{(t)}_{kj}$. Specifically, we subtract the mean of the excess value and redistribute it proportionally to the other sibling nodes. This update maintains the sum of probabilities under each node equal to 1 while ensuring different nodes' probabilities are adjusted proportionally. 

The rebalancing strategy is applied as follows if $ p^{(t)}_{ij} > \sum_{k \in U_{ij}} p^{(t)}_{kj}$. Let $d = p^{(t)}_{ij} - \sum_{k \in U_{ij}} p^{(t)}_{kj}$.

\begin{equation*}
\label{eq:1}
\begin{cases}
\begin{aligned}
p^{(t)}_{ij} & = p^{(t)}_{ij} - d 
\end{aligned} \\
p^{(t)}_{kj}  = p^{(t)}_{kj}  + \frac{d}{|U_{ij}|}, \text{for $k \in U_{ij}$}
\end{cases}
\end{equation*}
Leveraging the methods as mentioned earlier, we give the adjustment algorithm, Adj\_\,Probs($C$, $mode$, $s$, $i$) (Algorithm~\ref{alg:adjust}). 

\begin{algorithm}[H]
\caption{Adj\_Probs($C$, $f()$, $mode$, $s$, $i$)}
\label{alg:adjust}
\begin{algorithmic}[1]
\REQUIRE Concept hierarchy $C$, Target model $f()$, $mode$: LiRA-ADI or Entropy-ADI, 
         Image $s$ is drawn from the leaf node $n_{iL}$
\ENSURE Updated concept hierarchy $C$

\STATE Initialize: $j \gets L$
\WHILE{$j \neq 1$}
    \STATE $p^{(t)}_{ij} \gets p^{(t-1)}_{ij} + w(s, mode)\,\delta(j)$
    \FORALL{$k \in U_{ij}$}
        \STATE $p^{(t)}_{ij} \gets p^{(t-1)}_{ij} 
               \;-\; w(s, mode)\,\frac{\delta(j)}{\left|U_{ij}\right|}$
    \ENDFOR
    \STATE \textbf{Rebalancing step}
    \STATE $j \gets j - 1$
    \STATE $i \gets$ \text{index of the parent node of the current node}
\ENDWHILE
\RETURN Updated concept hierarchy $C$
\end{algorithmic}
\end{algorithm}

\subsubsection{Convergence Condition}\label{sec:converge}
Once the relevant concepts get boosted probabilities, the late iterations will more likely fetch the samples relevant to target domains. This positive feedback continues until convergence. A critical question is when we should stop the iterations. With the rebalancing strategy, the relevant node's probability eventually converges to $ p^{(t)}_{ij} \approx \sum_{k \in U_{ij}} p^{(t)}_{kj}$. However, it's not sufficient to serve as a global convergence condition. 

We design a global strategy as follows. The intuition is that if the majority of the sampled batch $I_t$ fits the target model $f()$ well, i.e., for most of them the normalized entropy $\tilde{H} (f(I_{tk})) \leq \lambda$ in Entropy-ADI, or the probability $\Lambda(f(I_{tk}))\leq 0.5$ in LiRA-ADI, we consider the concept probability adjustment converges. Thus, the converge function is a Boolean function defined as follows.

\begin{align}
    converge_{entropy}(I_t) = & \frac{1}{b} \sum_k \tilde{H}(I_{tk}) \leq \lambda ? 1 : 0 \\
    \text{OR} \nonumber \\
    converge_{LiRA}(I_t) = & \frac{1}{b} \sum_k \Lambda(I_{tk}) \geq 0.5 ? 1 : 0 
\end{align}
 
In the entropy-based method, we have carefully evaluated the setting of $\lambda$ in experiments and found that $\lambda=0.83$ works best with different experimental datasets.

It's important to note that the convergence will fail when the concept hierarchy is not large enough to cover any class of the target domain. As observed in experiments, most convergences happen around 50 iterations. We may use a relaxed upper bound, e.g., 100 iterations, to determine whether convergence is not possible, i.e., the concept hierarchy may not contain any class of the target domain.

\section{Experiments}
\label{sec:experiments}
\begin{figure*}[h!]
    \centering
    \begin{subfigure}[b]{0.24\textwidth}
        \centering
        \begin{tikzpicture}[scale=0.5]
            \tikzset{every picture/.style={font=\huge}} % Set all text to small
            \begin{axis}[
                ybar,
                bar width=.25cm,
                enlargelimits=0.25,
                legend style={at={(0.5,1.33)}, anchor=north, legend columns=2, yshift=-5pt},
                symbolic x coords={NETTE, Woof, Fashion},
                xtick=data,
                ylabel={OTDD},
                xlabel={Target domain},
                x tick label style={rotate=45, anchor=east},
                xlabel style={yshift=-24pt},
                ylabel style={font=\huge},
                tick label style={font=\huge},
                every node near coord/.append style={yshift=2pt}
                ]
                \addplot+[error bars/.cd, y dir=both, y explicit]
                    table[x=Datasets, y=ADI, y error=ADI_std, col sep=comma] {results/OTDD_high.csv};
                \addplot+[error bars/.cd, y dir=both, y explicit]
                    table[x=Datasets, y=ADI_E, y error=ADI_std_E, col sep=comma] {results/OTDD_high.csv};
                \addplot+[error bars/.cd, y dir=both, y explicit]
                    table[x=Datasets, y=HDI, y error=HDI_std, col sep=comma] {results/OTDD_high.csv};
                \addplot+[error bars/.cd, y dir=both, y explicit]
                    table[x=Datasets, y=GDI, y error=GDI_std, col sep=comma] {results/OTDD_high.csv};
                \legend{LiRA\_ADI, Entropy\_ADI, LDI, GDI}
            \end{axis}
        \end{tikzpicture}
        \caption{Similarity performance.}
        \label{exp:otdd}
    \end{subfigure}
    \hfill
    \begin{subfigure}[b]{0.24\textwidth}
        \centering
        \begin{tikzpicture}[scale=0.5]
            \tikzset{every picture/.style={font=\huge}} % Set all text to small
            \begin{axis}[
                ybar,
                bar width=.25cm,
                enlargelimits=0.15,
                legend style={at={(0.5,1.33)}, anchor=north, legend columns=3, yshift=-5pt},
                symbolic x coords={NETTE, Woof, Fashion},
                xtick=data,
                ylabel={ACC},
                xlabel={Target domain},
                x tick label style={rotate=45, anchor=east},
                xlabel style={yshift=-24pt},
                ylabel style={font=\huge},
                tick label style={font=\huge},
                every node near coord/.append style={yshift=2pt}
                ]
                \addplot+[error bars/.cd, y dir=both, y explicit]
                    table[x=Datasets, y=ADI, y error=std_HDI, col sep=comma] {results/MI_high.csv};
                \addplot+[error bars/.cd, y dir=both, y explicit]
                    table[x=Datasets, y=ADI_E, y error=std_ADI_E, col sep=comma] {results/MI_high.csv};
                \addplot+[error bars/.cd, y dir=both, y explicit]
                    table[x=Datasets, y=HDI, y error=std_HDI, col sep=comma] {results/MI_high.csv};
                \addplot+[error bars/.cd, y dir=both, y explicit]
                    table[x=Datasets, y=GDI, y error=std_GDI, col sep=comma] {results/MI_high.csv};
                                 \addplot+[error bars/.cd, y dir=both, y explicit]
                    table[x=Datasets, y=Baseline, y error=std_baseline, col sep=comma] {results/MI_high.csv};
                \legend{ LiRA\_ADI, Entropy\_ADI, LDI, GDI,Baseline}
            \end{axis}
        \end{tikzpicture}
        \caption{Model inversion enhancement.}
        \label{exp:MI}
    \end{subfigure}
    \hfill
\begin{subfigure}[b]{0.38\textwidth}
    \centering
    \begin{tikzpicture}[scale=0.5]
        \tikzset{every picture/.style={font=\huge}} % Set all text to large
        \begin{axis}[
            ybar,
            bar width=.3cm,
            enlargelimits=0.15,
            legend style={at={(0.5,1.33)}, anchor=north, legend columns=2, yshift=-5pt},
            symbolic x coords={NETTE, Woof, Fashion},
            xtick=data,
            ylabel={\# of Accesses},
            xlabel={Target domain},
            x tick label style={rotate=45, anchor=east},
            xlabel style={yshift=-24pt},
            ylabel style={font=\huge},
            tick label style={font=\huge},
            clip=false
            ]
            \addplot+[error bars/.cd, y dir=both, y explicit]
                table[x=Datasets, y=ADI, y error=ADI_std, col sep=comma] {results/access_high.csv};
            \addplot+[error bars/.cd, y dir=both, y explicit]
                table[x=Datasets, y=ADI_E, y error=ADI_std_E, col sep=comma] {results/access_high.csv};
            \addplot+[error bars/.cd, y dir=both, y explicit]
                table[x=Datasets, y=HDI, col sep=comma] {results/access_high.csv};
            \addplot+[error bars/.cd, y dir=both, y explicit]
                table[x=Datasets, y=GDI, y error=GDI_std, col sep=comma] {results/access_high.csv};
            \legend{LiRA\_ADI, Entropy\_ADI, LDI, GDI}

            % Draw the magnified rectangle
            \draw[thick, dashed, red] 
    ([shift={(-27,-20)}] axis cs:NETTE, 0) rectangle ([shift={(-100,-20)}] axis cs:Woof, 5000);

                  % Add connecting lines to zoomed plot
            \draw[thick, red, ->]  ([shift={(-27,0)}] axis cs:NETTE, 0) -- ([shift={(-180,-20)}] axis cs:Woof, 5000);
         
        \end{axis}

        % Add zoomed-in plot as a node
 \node[anchor=north east, xshift=-3.7cm, yshift=-1.3cm] at (rel axis cs:1,1) {
    \begin{tikzpicture}
        \begin{axis}[
            width=2.5cm,
            height=2.5cm,
            ybar,
            bar width=0.15cm,
            enlargelimits=0.15,
            xtick=data, % Fix for repeated labels
            symbolic x coords={NETTE}, % Define symbolic x-coordinates
            xticklabel style={font=\small, rotate=45, anchor=east},
            tick label style={font=\small},
            ymin=0, ymax=100,
            ]
            \addplot+[error bars/.cd, y dir=both, y explicit]
                table[x=Datasets, y=ADI, col sep=comma] {results/access_high_2.csv};
            \addplot+[error bars/.cd, y dir=both, y explicit]
                table[x=Datasets, y=ADI_E, col sep=comma] {results/access_high_2.csv};
        \end{axis}
    \end{tikzpicture}
};
    \end{tikzpicture}
     \caption{Number of accesses with magnifier effect.}
    \label{exp:accesses}
\end{subfigure}
    \caption{Comparison of performance on high-resolution datasets and large-scale concept hierarchy. ADI outperforms other domain inference attacks in both efficiency and efficacy.}
    \label{fig:OTDD_mix_barplot}
\end{figure*}

ADI aims to address the limitations of the GAN-based GDI method \cite{gu2022}: it's difficult to identify candidate datasets at the dataset level similar to the target training data, and excessive target model accesses are also disadvantageous in private model application scenarios, e.g., the attack is more likely to be detected. Our experiments demonstrate that ADI can effectively mitigate these limitations. Specifically, the experiments will achieve the following goals. (1) We demonstrate that our attacks provide more accurate domain estimations compared to GDI on high-resolution, ImageNet-related datasets and complex concept hierarchies. (2) Our attacks require significantly fewer accesses to the target model to achieve superior results, which is crucial for attacks in privacy-sensitive environments. (3) We have thoroughly evaluated the impact of various parameter settings on ADI's performance.

\subsection{Setup}
\label{sec:setup}
\textbf{Concept hierarchies and datasets.} We tested three methods for creating concept hierarchies. 
(1) We used a dataset-based hierarchy and synthesized domain datasets for parameter tuning. The three-layer hierarchy includes MNIST \cite{MNIST}, EMNIST \cite{EMNIST}, LFW \cite{LFW}, CIFAR10, CIFAR100 \cite{CIFAR}, CINIC-10 \cite{CINIC}, FASHION-MNIST \cite{FASHION}, and CLOTHING \cite{Clothing}. Each dataset is an intermediate node, with classes as leaf concepts. Images are scaled to 36x36x3. For mixed-domain training, we randomly selected 10, 20, and 30 classes from the 8 datasets to build synthesized training sets. Unselected data formed the adversaries' data pool and hierarchy.

(2) We also experimented with a larger hierarchy from ImageNet-1k and three ImageNet-related datasets: ImageNETTE \cite{imagenette}, ImageWoof \cite{imagenette}, and DeepFashion \cite{liu2016deepfashion}. The hierarchy includes 1000 leaf nodes (ImageNet-1k classes) and up to 10 intermediate layers\footnote{The ImageNet-1K hierarchy is available at our code link.}. DeepFashion uses 10 classes from the "Clothing" branch of ImageNet. ImageNETTE includes 10 easily classified ImageNet classes, and ImageWoof includes dog breeds from ImageNet. These examples demonstrate the functionality of a larger hierarchy. For more details, see Appendix C.

\textbf{Target models.} We used ResNet-18 \cite{he2016} for the three synthesized datasets, training with an 8:2 split, repeated 10 times to observe variance. The training used SGD with a learning rate of $10^{-2}$, batch size of 100, momentum of 0.9, and learning rate decay of $10^{-4}$.

For the three high-resolution ImageNet-related datasets, we used ResNet-152, with the same split, repetitions, and hyperparameters as the ResNet-18 models. All models were trained using the Fast Forward Computer Vision (FFCV) pipeline for efficiency \cite{FFCV}.

\textbf{Shadow models.} Shadow models for depicting out-domain samples’ output confidence vectors are crucial for the hypothesis-based membership inference attack and, thus, vital to both LDI and ADI. Since class numbers vary, we separate the data pool by target domain classes and train shadow models accordingly. For a $k$-class target domain, we divide the attack pool into multiple $k$-class sub-domains and randomly sample $n$ sub-domains to train $n$ shadow models per sub-domain. We set $n=128$ for synthesized datasets and $n=256$ for ImageNet-related datasets, as suggested in \cite{carlini2022lira}. For details on shadow models' impacts, see Appendix B.

\subsection{Evaluation Metrics}
\label{evaluation metrics}

\textbf{Dataset similarity.} As mentioned in Section \ref{preliminery}, we employ the Optimal Transport Dataset Distance (OTDD) to quantify dataset similarity between the original target dataset and the dataset extracted by the attack. Lower OTDD values indicate greater similarity between the estimated dataset and the original domain.

\textbf{Performance of model inversion attack.} The GDI, LDI, and ADI estimated data will be used as the auxiliary datasets to the model-inversion attack -- if the estimated data is similar to the original domain, the accuracy of the model-inversion attack will be significantly boosted. We use the original model-inversion attack designed by Fredrikson et al. \cite{fredrikson2015}, the effectiveness of which is evaluated by the quality of its reconstructed training dataset. We use the target model to recognize (i.e., classify) the reconstructed images -- the higher the classification accuracy, the better the reconstructed data and thus the better the estimated data.

\subsection{Results on ImageNet-Related Datasets} 
\label{sec:access_exp}
\begin{figure}[ht!]
    \centering
    \begin{subfigure}[b]{0.23\textwidth}
        \centering
        \begin{tikzpicture}
        \tikzset{every picture/.style={font=\small}}
        \begin{axis}[
            width=\textwidth,
            height=0.8\textwidth,
            xlabel={Number of accesses (\%)},
            ylabel={OTDD},
            legend style={at={(1.3,1.05)}, font=\small, anchor=south, legend columns=-1},
            grid=both
        ]
        \addplot+[
            error bars/.cd,
            y dir=both, y explicit
        ] coordinates {
            (20, 232.54) 
            (40, 225.63) 
            (60, 201.42) 
            (80, 183.23) 
            (100, 156.23) 
        };
        \addlegendentry{Mix-10}

        \addplot+[
            error bars/.cd,
            y dir=both, y explicit
        ] coordinates {
            (20, 242.42) 
            (40, 228.21)
            (60, 217.43)
            (80, 197.43) 
            (100, 185.32)
        };
        \addlegendentry{Mix-20}

        \addplot+[
            error bars/.cd,
            y dir=both, y explicit
        ] coordinates {
            (20, 247.31)
            (40, 242.54)
            (60, 231.46)
            (80, 215.36)
            (100, 195.35)
        };
        \addlegendentry{Mix-30}
        \end{axis}
        \end{tikzpicture}
        \caption{Synthesized sets(balanced)}
    \end{subfigure}
     \hfill
         \begin{subfigure}[b]{0.23\textwidth}
        \centering
        \begin{tikzpicture}
        \tikzset{every picture/.style={font=\small}}
        \begin{axis}[
            width=\textwidth,
            height=0.8\textwidth,
            xlabel={Number of accesses (\%)},
            legend style={at={(0.5,1.05)}, font=\small, anchor=south, legend columns=-1},
            grid=both
        ]
        \addplot+[
            error bars/.cd,
            y dir=both, y explicit
        ] coordinates {
            (20, 239.23) 
            (40, 215.54) 
            (60, 198.83) 
            (80, 192.32) 
            (100, 164.44) 
        };
        \addplot+[
            error bars/.cd,
            y dir=both, y explicit
        ] coordinates {
            (20, 251.33) 
            (40, 232.45)
            (60, 221.15)
            (80, 204.32) 
            (100, 178.56)
        };
        \addplot+[
            error bars/.cd,
            y dir=both, y explicit
        ] coordinates {
            (20, 254.16)
            (40, 236.53)
            (60, 213.66)
            (80, 198.25)
            (100, 187.54)
        };
        \end{axis}
        \end{tikzpicture}
         
        \caption{Synthesized sets(random)}
    \end{subfigure}
     \hfill
    \begin{subfigure}[b]{0.23\textwidth}
        \centering
        \begin{tikzpicture}
        \tikzset{every picture/.style={font=\small}}
        \begin{axis}[
            width=\textwidth,
            height=0.8\textwidth,
            legend style={at={(1.3,1.05)}, font=\small, anchor=south, legend columns=4},
            ylabel={OTDD},
            xlabel={Number of accesses (\%)},
            grid=both
        ]
        \addplot+[
            error bars/.cd,
            y dir=both, y explicit
        ] coordinates {
            (20, 167.53) 
            (40, 156.32) 
            (60, 131.25) 
            (80, 127.45) 
            (100, 115.32) 
        };
        \addlegendentry{M}

        \addplot+[
            error bars/.cd,
            y dir=both, y explicit
        ] coordinates {
            (20, 184.32) 
            (40, 182.24) 
            (60, 169.34)
            (80, 153.33) 
            (100, 124.76) 
        };
        \addlegendentry{E}

        \addplot+[
            error bars/.cd,
            y dir=both, y explicit
        ] coordinates {
            (20, 175.74) 
            (40, 163.24) 
            (60, 138.53) 
            (80, 125.46) 
            (100, 95.35) 
        };
        \addlegendentry{L}

        \addplot+[
            error bars/.cd,
            y dir=both, y explicit
        ] coordinates {
            (20, 192.43) 
            (40, 175.74) 
            (60, 172.53) 
            (80, 164.23) 
            (100, 132.53) 
        };
        \addlegendentry{C-10}

        \addplot+[
            error bars/.cd,
            y dir=both, y explicit
        ] coordinates {
            (20, 232.65) 
            (40, 201.54) 
            (60, 182.14) 
            (80, 168.94) 
            (100, 152.32) 
        };
        \addlegendentry{C-100}

        \addplot+[
            error bars/.cd,
            y dir=both, y explicit
        ] coordinates {
            (20, 215.35) 
            (40, 203.25) 
            (60, 183.22) 
            (80, 164.36) 
            (100, 147.32) 
        };
        \addlegendentry{CN-10}

        \addplot+[
            error bars/.cd,
            y dir=both, y explicit
        ] coordinates {
            (20, 192.65) 
            (40, 179.75) 
            (60, 172.35)
            (80, 154.33) 
            (100, 148.54) 
        };
        \addlegendentry{FM}

        \addplot+[
            error bars/.cd,
            y dir=both, y explicit
        ] coordinates {
            (20, 226.54) 
            (40, 214.32) 
            (60, 181.26) 
            (80, 173.43) 
            (100, 156.43)
        };
        \addlegendentry{CL}
        \end{axis}
        \end{tikzpicture}
        \caption{Single sets(balanced)}
    \end{subfigure}
         \hfill
    \begin{subfigure}[b]{0.23\textwidth}
        \centering
        \begin{tikzpicture}
        \tikzset{every picture/.style={font=\small}}
        \begin{axis}[
            width=\textwidth,
            height=0.8\textwidth,
            legend style={at={(0.5,1.05)}, font=\small, anchor=south, legend columns=4},
            xlabel={Number of accesses (\%)},
            grid=both
        ]
        \addplot+[
            error bars/.cd,
            y dir=both, y explicit
        ] coordinates {
            (20, 182.32) 
            (40, 162.54) 
            (60, 142.35) 
            (80, 132.53) 
            (100, 115.32) 
        };
        \addplot+[
            error bars/.cd,
            y dir=both, y explicit
        ] coordinates {
            (20, 179.24) 
            (40, 168.37) 
            (60, 166.25)
            (80, 154.46) 
            (100, 131.25) 
        };
        \addplot+[
            error bars/.cd,
            y dir=both, y explicit
        ] coordinates {
            (20, 181.57) 
            (40, 174.53) 
            (60, 145.42) 
            (80, 136.21) 
            (100, 97.25) 
        };

        \addplot+[
            error bars/.cd,
            y dir=both, y explicit
        ] coordinates {
            (20, 195.22) 
            (40, 183.31) 
            (60, 165.25) 
            (80, 157.53) 
            (100, 129.31) 
        };

        \addplot+[
            error bars/.cd,
            y dir=both, y explicit
        ] coordinates {
            (20, 223.56) 
            (40, 196.27) 
            (60, 189.46) 
            (80, 174.94) 
            (100, 154.25) 
        };

        \addplot+[
            error bars/.cd,
            y dir=both, y explicit
        ] coordinates {
            (20, 231.26) 
            (40, 225.28) 
            (60, 193.74) 
            (80, 191.21) 
            (100, 146.22) 
        };

        \addplot+[
            error bars/.cd,
            y dir=both, y explicit
        ] coordinates {
            (20, 201.36) 
            (40, 185.38) 
            (60, 164.75)
            (80, 159.85) 
            (100, 146.25) 
        };

        \addplot+[
            error bars/.cd,
            y dir=both, y explicit
        ] coordinates {
            (20, 234.74) 
            (40, 221.75) 
            (60, 197.15) 
            (80, 183.93) 
            (100, 156.35)
        };
        \end{axis}
        \end{tikzpicture}
        \caption{Single sets(random)}
    \end{subfigure}
     \caption{OTDD constantly decreases with the increasing size of tested samples in LDI. Dataset Names: M-MNIST, E-EMNIST, L-LFW, C-CIFAR, CN-CINIC, FM-FashionMNIST, CL-CLOTHING, NETTE-ImageNETTE, Woof-ImageWoof, Fashion-DeepFashion.}
    \label{fig:Advanced_LDI}
\end{figure}

We show the methods' attacking performance on the three ImageNet-related datasets. Due to page limitations, we include the results on synthesized datasets in Appendix D.
We use the parameters that generate the best-estimated datasets with the smallest OTDD scores for the attacking methods. For LDI, we select 100\% of instances from balanced classes. For ADI, we use the optimal parameter settings for $\lambda$ and $\delta(j)$. We will discuss the parameter settings in later sections. 

\begin{figure*}[h!]
    \centering
    \begin{subfigure}{0.32\textwidth}
        \centering
        \begin{tikzpicture}[scale=0.4]
        \tikzset{every picture/.style={font=\Huge}} % Set all text to small
        \begin{axis}[
            xlabel={Epochs},
            ylabel={OTDD},
            ylabel style={xshift=-10pt},
            grid=major,
            ymin=0,
            ymax=500,
            legend style={at={(0.5,1.35)}, anchor=north, legend columns=3, font=\Huge},
        ]
        \addplot+[mark=*, mark size=0.5pt, color=black!60!green, dashed, error bars/.cd, y dir=both, y explicit] 
        table [x=Epochs, y=OTDD, col sep=comma] {results/Mix-10.csv};

        \addplot+[mark=*, mark size=0.5pt, color=red!80, dashed, error bars/.cd, y dir=both, y explicit] 
        table [x=Epochs, y=OTDD, col sep=comma] {results/Mix-20.csv};

        \addplot+[mark=*, mark size=0.5pt, color=blue!70!green, dashed, error bars/.cd, y dir=both, y explicit] 
        table [x=Epochs, y=OTDD, col sep=comma] {results/Mix-30.csv};

        \addplot+[mark=*, mark size=0.5pt, color=black!60!green, solid, error bars/.cd, y dir=both, y explicit] 
        table [x=Epochs, y=OTDD, col sep=comma] {results/Mix-10L.csv};

        \addplot+[mark=*, mark size=0.5pt, color=red!80, solid, error bars/.cd, y dir=both, y explicit] 
        table [x=Epochs, y=OTDD, col sep=comma] {results/Mix-20L.csv};

        \addplot+[mark=*, mark size=0.5pt, color=blue!70!green, solid, error bars/.cd, y dir=both, y explicit] 
        table [x=Epochs, y=OTDD, col sep=comma] {results/Mix-30L.csv};
  
        \end{axis}
        \end{tikzpicture}
        \caption{Setting $\delta = \frac{j}{|C|}$: Proper $\delta$ makes quick convergence.}
        \label{fig:proper delta}
    \end{subfigure}
    \hfill
    \begin{subfigure}{0.32\textwidth}
        \centering
        \begin{tikzpicture}[scale=0.4]
        \tikzset{every picture/.style={font=\Huge}} % Set all text to small
        \begin{axis}[
            xlabel={Epochs},
            ylabel={OTDD},
            grid=major,
            ymin=0,
            ymax=500,
            legend style={at={(0.5,1.20)}, anchor=north, legend columns=3, font=\Huge},
        ]
        \addplot+[mark=*, mark size=0.5pt, color=black!60!green, dashed, error bars/.cd, y dir=both, y explicit] 
        table [x=Epochs, y=OTDD, col sep=comma] {results/Mix-10-2.csv};
        \addlegendentry{Mix-10-entropy}

        \addplot+[mark=*, mark size=0.5pt, color=red!80, dashed, error bars/.cd, y dir=both, y explicit] 
        table [x=Epochs, y=OTDD, col sep=comma] {results/Mix-20-2.csv};
        \addlegendentry{Mix-20-entropy}

        \addplot+[mark=*, mark size=0.5pt, color=blue!70!green, dashed, error bars/.cd, y dir=both, y explicit] 
        table [x=Epochs, y=OTDD, col sep=comma] {results/Mix-30-2.csv};
        \addlegendentry{Mix-30-entropy}

        \addplot+[mark=*, mark size=0.5pt, color=black!60!green, solid, error bars/.cd, y dir=both, y explicit] 
        table [x=Epochs, y=OTDD, col sep=comma] {results/Mix-10-2L.csv};
        \addlegendentry{Mix-10-LiRA}

        \addplot+[mark=*, mark size=0.5pt, color=red!80, solid, error bars/.cd, y dir=both, y explicit] 
        table [x=Epochs, y=OTDD, col sep=comma] {results/Mix-20-2L.csv};
        \addlegendentry{Mix-20-LiRA}

        \addplot+[mark=*, mark size=0.5pt, color=blue!70!green, solid, error bars/.cd, y dir=both, y explicit] 
        table [x=Epochs, y=OTDD, col sep=comma] {results/Mix-30-2L.csv};
        \addlegendentry{Mix-30-LiRA}
        \end{axis}
        \end{tikzpicture}
        \caption{Setting $\delta = \frac{10j}{|C|}$: Large $\delta(j)$ causes oscillation, difficult to converge.}
        \label{fig:greater delta}
    \end{subfigure}
    \hfill
    \begin{subfigure}{0.32\textwidth}
        \centering
        \begin{tikzpicture}[scale=0.4]
        \tikzset{every picture/.style={font=\Huge}} 
        \begin{axis}[
            xlabel={Epochs},
            ylabel={OTDD},
            grid=major,
            ymin=0,
            ymax=500,
             legend style={at={(0.5,1.35)}, anchor=north, legend columns=3, font=\Huge},
        ]
        \addplot+[mark=*, mark size=0.5pt, color=black!60!green, dashed, error bars/.cd, y dir=both, y explicit] 
        table [x=Epochs, y=OTDD, col sep=comma] {results/Mix-10-3.csv};

        \addplot+[mark=*, mark size=0.5pt, color=red!80, dashed, error bars/.cd, y dir=both, y explicit] 
        table [x=Epochs, y=OTDD, col sep=comma] {results/Mix-20-3.csv};

        \addplot+[mark=*, mark size=0.5pt, color=blue!70!green, dashed, error bars/.cd, y dir=both, y explicit] 
        table [x=Epochs, y=OTDD, col sep=comma] {results/Mix-30-3.csv};

        \addplot+[mark=*, mark size=0.5pt, color=black!60!green, solid, error bars/.cd, y dir=both, y explicit] 
        table [x=Epochs, y=OTDD, col sep=comma] {results/Mix-10-3L.csv};

        \addplot+[mark=*, mark size=0.5pt, color=red!80, solid, error bars/.cd, y dir=both, y explicit] 
        table [x=Epochs, y=OTDD, col sep=comma] {results/Mix-20-3L.csv};

        \addplot+[mark=*, mark size=0.5pt, color=blue!70!green, solid, error bars/.cd, y dir=both, y explicit] 
        table [x=Epochs, y=OTDD, col sep=comma] {results/Mix-30-3L.csv};

        \end{axis}
        \end{tikzpicture}
        \caption{Setting $\delta = \frac{j}{10|C|}$: Lower $\delta$ slows convergence.}
        \label{fig:smaller delta}
    \end{subfigure}
    \caption{Examining the effects of $\delta$ configurations.}
\end{figure*}

\textbf{Quality of estimated domain.} To observe GDI's performance, we split ImageNet-1K into 100 datasets with 10 similar classes each to build the landmark datasets, such as 10 dog breeds, 10 types of clothing, etc. The results in Figure \ref{exp:otdd} show that LiRA-ADI performs the best with the smallest OTDD score, and Entropy-ADI also outperforms LDI and GDI.

\textbf{Enhancing model inversion attacks.} Model inversion attacks require knowledge of the target domain to implement or enhance the attack. We are also interested in how the estimated domain can enhance model inversion attacks. We recover 1000 images for each class of the target model using model inversion attacks and test the target model's performance on these recovered datasets. In the baseline scenario, we initialize the generated images randomly. In other experiments, we randomly pick images from the estimated domain as the initial images. Figure \ref{exp:MI} shows the enhancement of model inversion attacks. Both ADI methods outperform other attacks with LDI ranked next. 

\textbf{Model Accesses.} One motivation for designing our attacks is to reduce the substantial number of accesses required by GDI. Figure \ref{exp:accesses} illustrates the number of accesses needed for each attack to reach convergence. We set the batch size for drawing instances and interacting with the target model is 1000. As shown in the results, ADI requires significantly fewer accesses to implement compared to GDI and LDI. The large number of accesses needed by GDI is due to the necessity of training a generative model for each dataset, which involves multiple interactions with the target model. LDI's model access is critically related to the result quality. We show in the next section that LDI needs to test almost all samples in the data pool to get the best-performing results, which can be expensive. 

\subsection{Parameter Settings}
\label{sec:parameters}
In this section, we include more details for determining the optimal parameter settings for LDI and ADI methods.

\subsubsection{Effect of Sampling Methods for LDI}
\label{sec:LDI_result}

To understand the effect of different sampling methods on the LDI results, we progressively increase the number of tested samples. For simplicity, we augmented small classes to make the class sizes even so that we can use percentages to represent the sampling progress for both uniform sampling and class-balanced sampling. The results are shown in Figure \ref{fig:Advanced_LDI}. We conducted experiments on two types of target domains (and target models): a synthesized target domain consisting of randomly selected classes in the dataset pool, denoted by Mix-10, Mix-20, and Mix-30; uniform samples from each dataset, denoted by the corresponding dataset.  The two sampling methods are used to generate the test samples, aiming to reduce the cost of LDI, denoted by ``random'' and ``balanced''. However, we found that the LDI's performance is almost linearly related to the number of test samples. It does not reach the peak performance until all samples are tested. This pattern keeps across all datasets and sampling methods, which raises the concern that to achieve the best LDI performance, the number of model accesses will be significantly high. However, even with such a high cost, LDI still performs 20\% worse than ADI.  

\subsubsection{ADI Parameter Settings}

The ADI algorithm contains several parameters to be experimentally explored and determined, including the threshold of the normalized entropy, $\lambda$ in the entropy-based method, and the layer-wise adaptive probability adjustment $\delta$ in both entropy and LiRA-based method. We also want to observe the effect of probability rebalancing and assess the benefits of using concept hierarchies against a flat concept list. 

\textbf{$\delta$ settings.} The layer-wise probability adjustment $\delta$ plays a crucial role in the speed, quality of convergence, and attack efficiency. We found that $\delta(j) = j/|C|$ for nodes at Level $j$ and the total number of concepts $|C|$ works reasonably well according to the heuristic mentioned in Section \ref{sec:concept adjustment}(Figure \ref{fig:proper delta}). We have also tested variants of $\delta$ settings, i.e., $\delta(j) = \frac{10j}{|C|}$ and $\delta=\frac{j}{10|C|}$. With $\delta(j) = \frac{10j}{|C|}$, we observed significant oscillations at higher OTDD levels, indicating the attack does not converge well (Figure \ref{fig:greater delta}). Conversely, reducing $\delta(j)$ to $\frac{j}{10|C|}$ resulted in a substantial slowdown in convergence, increasing the attack cost (Figure \ref{fig:smaller delta}).

\textbf{Effect of rebalancing.} Without probability rebalancing, ADI may lead to a biased concept distribution, i.e., focusing on increasing the probabilities of a few branches of the concept hierarchy due to their already high probabilities. Other methods, such as fine-tuning the $\delta$ values, may help address this issue. However, we find the rebalancing method works extremely well. Experiments show a remarkable improvement in ADI performance with rebalancing, as shown in Figure \ref{fig:punishment}. Without rebalancing, the ADI gives unstable results. Due to the biased adjustment towards certain branches, the results are often stuck at suboptimal levels. 
\begin{figure}[ht!]
    \centering
    \begin{subfigure}[b]{0.23\textwidth}
        \centering
        \begin{tikzpicture}[scale=0.4]
        \tikzset{every picture/.style={font=\Huge}} 
            \begin{axis}[
                ybar,
                bar width=6pt,
                symbolic x coords={Mix-10, Mix-20, Mix-30},
                xtick=data,
                ylabel={OTDD},
                xlabel={Dataset},
                ymajorgrids=true,
                grid style=dashed,
                xlabel style={yshift=-15pt},
                legend style={at={(1.35,1.15)},anchor=north,font=\Huge},
                legend columns=-1,
                transpose legend,
            ]
            \addplot+[error bars/.cd, y dir=both, y explicit] table [x=Dataset, y=OTDD_without, y error=std_without, col sep=comma] {results/MI_punish.csv};
            \addlegendentry{w/out rebalancing}
            \addplot+[error bars/.cd, y dir=both, y explicit] table [x=Dataset, y=OTDD_with, y error=std_with, col sep=comma] {results/MI_punish.csv};
            \addlegendentry{with rebalancing}
            \end{axis}
        \end{tikzpicture}   
        \caption{Entropy-based ADI}
        \label{fig:entropy}
    \end{subfigure}
    \hfill
    \begin{subfigure}[b]{0.23\textwidth}
        \centering
        \begin{tikzpicture}[scale=0.4]
        \tikzset{every picture/.style={font=\Huge}} 
            \begin{axis}[
                ybar,
                bar width=6pt,
                symbolic x coords={Mix-10, Mix-20, Mix-30},
                xtick=data,
                ylabel={OTDD},
                xlabel={Dataset},
                xlabel style={yshift=-15pt},
                ymajorgrids=true,
                grid style=dashed,
                legend style={at={(0.5,1.15)},anchor=north,font=\Huge},
                legend columns=-1,
                transpose legend,
            ]
            \addplot+[error bars/.cd, y dir=both, y explicit] table [x=Dataset, y=OTDD_without, y error=std_without, col sep=comma] {results/MI_punish_L.csv};
            \addplot+[error bars/.cd, y dir=both, y explicit] table [x=Dataset, y=OTDD_with, y error=std_with, col sep=comma] {results/MI_punish_L.csv};
            \end{axis}
        \end{tikzpicture}
         
        \caption{LiRA-based ADI}
        \label{fig:LiRA}
    \end{subfigure}
     
    \caption{Comparison of entropy-based and LiRA-based methods with and without rebalancing on different datasets.}
    \label{fig:punishment}
\end{figure}

\textbf{Effect of Concept Hierarchy.} We wonder how the multi-layer hierarchical structure may affect the ADI algorithm, compared to a flat list of concepts. We construct the flat structure by removing all the internal nodes of the concept tree, i.e., the root node pointing to all leaves directly, so that the ADI algorithm still works without modification. The algorithm still converges, but with much worse quality as shown in Figure \ref{fig:OTDD_C_Hierarchy}. Without the hierarchically organized concepts, the ADI estimated domains yield larger OTDD values, meaning less similar to the desired domain distributions, and the variances are also larger. The result indicates that the intermediate nodes can direct the probability adjustments more to the focal concepts, resulting in a better-estimated concept distribution. In contrast, the flat concept structure disperses the probability adjustments.  

\begin{figure}[h!]
    \centering
    \begin{subfigure}[b]{0.23\textwidth}
        \centering
        \begin{tikzpicture}[scale=0.4]
        \tikzset{every picture/.style={font=\Huge}} 
            \begin{axis}[
                xlabel={Epochs},
                ylabel={OTDD},
                grid=major,
                ymin=80,
                ymax=500,
                legend style={font=\Huge, at={(1.35,1.23)}, anchor=north, legend columns=-1, yshift=-10pt},
            ]  
            \addplot+[mark size=0.5pt,solid,color=black!60!green,error bars/.cd,y dir=both,y explicit] table [x=Epochs, y=OTDD_E, col sep=comma] {results/Mix-10-flatten.csv};
            \addlegendentry{Mix-10}
            \addplot+[mark size=0.5pt,solid,color=red!80,error bars/.cd,y dir=both,y explicit] table [x=Epochs, y=OTDD_E, col sep=comma] {results/Mix-20-flatten.csv};
            \addlegendentry{Mix-20}
            \addplot+[mark size=0.5pt,solid,color=blue!70!green,error bars/.cd,y dir=both,y explicit] table [x=Epochs, y=OTDD_E, col sep=comma] {results/Mix-30-flatten.csv};
            \addlegendentry{Mix-30}
            \end{axis}
        \end{tikzpicture}
        \caption{Entropy-based ADI}
        \label{fig:OTDD_Entropy}
    \end{subfigure}
    \hfill
    \begin{subfigure}[b]{0.23\textwidth}
        \centering
        \begin{tikzpicture}[scale=0.4]
        \tikzset{every picture/.style={font=\Huge}} 
            \begin{axis}[
                xlabel={Epochs},
                ylabel={OTDD},
                grid=major,
                ymin=80,
                ymax=500,
                legend style={font=\Huge, at={(0.5,1.2)}, anchor=north, legend columns=1, yshift=-10pt},
            ]  
            \addplot+[mark size=0.5pt,dotted,color=black!60!green,error bars/.cd,y dir=both,y explicit] table [x=Epochs, y=OTDD_L, col sep=comma] {results/Mix-10-flatten.csv};
           
            \addplot+[mark size=0.5pt,dotted,color=red!80,error bars/.cd,y dir=both,y explicit] table [x=Epochs, y=OTDD_L, col sep=comma] {results/Mix-20-flatten.csv};
     
            \addplot+[mark size=0.5pt,dotted,color=blue!70!green,error bars/.cd,y dir=both,y explicit] table [x=Epochs, y=OTDD_L, col sep=comma] {results/Mix-30-flatten.csv};

            \end{axis}
        \end{tikzpicture}
        \caption{LiRA-based ADI}
        \label{fig:OTDD_LiRA}
    \end{subfigure}
    \caption{OTDD values over epochs for ADI with flattened concept hierarchy.}
    \label{fig:OTDD_C_Hierarchy}
\end{figure}

\textbf{$\lambda$ settings in entropy-based ADI.} The $\lambda$ setting serves as the threshold, indicating (1) the target model predicts the extracted sample with high confidence, i.e., a positive sample, and (2) the overall quality of the batch of extracted samples, i.e., the convergence condition. Its setting is critical for the entropy-based ADI attack to converge quickly toward a high-quality result. We have conducted a set of experiments to investigate the optimal setting. Figure \ref{fig:lambda} shows variable settings of $\lambda$ and $\lambda = 0.83$ gives the best-extracted datasets among others. It's worthy noting that LiRA-based ADI does not need this parameter, which makes it more stable in practice.

\begin{figure}[h!]
\centering
\begin{tikzpicture}[scale=0.4]
\tikzset{every picture/.style={font=\Huge}} 
\begin{axis}[
xlabel={$\lambda$},
ylabel={OTDD},
xtick={0.77, 0.80, 0.83, 0.86, 0.89},
ymajorgrids=true,
grid style=dashed,
legend style={at={(0.5,1.15)},anchor=north,font=\huge},
legend columns=-1, % split legend into three columns
transpose legend,
]
\addplot+[error bars/.cd, y dir=both, y explicit] table [x=lambda, y=Mix-10, y error=Mix-10_std, col sep=comma] {results/lambda.csv};
\addlegendentry{Mix-10}

\addplot+[error bars/.cd, y dir=both, y explicit] table [x=lambda, y=Mix-20, y error=Mix-20_std, col sep=comma] {results/lambda.csv};
\addlegendentry{Mix-20}

\addplot+[error bars/.cd, y dir=both, y explicit] table [x=lambda, y=Mix-30, y error=Mix-30_std, col sep=comma] {results/lambda.csv};
\addlegendentry{Mix-30}

\end{axis}
\end{tikzpicture}
 
\caption{The setting of $\lambda$ significantly impacts the quality of the attack. $\lambda=0.83$ yields the best outcomes.}
\label{fig:lambda}
\end{figure}

\section{Related Work}
\label{sec:related work}
Machine learning models in fields like intrusion detection and healthcare are increasingly exposed to cyber threats through API services \cite{shameer2017,romero2021,alkhalil2021,gupta2017,pingle2018,al2020,bhushan2017}. Significant threats include model inversion and inference attacks, which compromise training data privacy. Model inversion attacks aim to recreate training examples using auxiliary data, with recent GAN-based methods achieving high-quality results \cite{zhang2020}. Membership inference attacks identify whether specific data was used in training \cite{shokri17, Hui21, Choquette-Choo21}, while property inference attacks reveal population-level attributes \cite{ganju2018,zhang2021}. Both strategies rely on understanding the target model's training data distribution. The Fidel attack on federated learning exploits neuron data to infer previous activations, posing a security risk that depends on having an auxiliary dataset resembling the victim's training data \cite{enthoven2022}.

While the assumption of an attacker knowing the auxiliary data or domain distribution may not hold for a breached private model API service, it is critical to investigate if accessing an unnamed model API only can also reveal domain information. A recent GAN-based model-domain inference (GDI) attack aims to address this \cite{gu2022}. However, GDI does not work satisfactorily when only a few parts of candidate datasets are relevant to the target dataset, and it requires excessive model accesses. Our proposed ADI attack directly addresses these two problems with the GDI attack.

\section{Conclusion}
\label{sec:conclusion}
Deep neural network models can be exploited by model-targeted attacks, e.g., model inversion and membership inference attacks. However, attackers cannot apply these attacks effectively without knowing the model's domain information. A possible protection approach is to package a sensitive model as an unnamed function call/API service, hiding the model's domain knowledge from attackers. However, we show that the Adaptive Domain Inference (ADI) attack can still identify the classes of samples relevant to the target model's domain by only accessing the unnamed model. By utilizing the concept hierarchy extracted from a large data pool, ADI's iterative procedure can efficiently and effectively tune the attack towards the concepts likely included by the target model’s training data. ADI surpasses the compared domain inference attacks GDI and LDI by a large margin in the extracted datasets' quality and significantly fewer model accesses. 

\textbf{Acknowledgement.} This material is based upon work supported by the National Science Foundation under Grant No. (2232824).

\bibliographystyle{ACM-Reference-Format}
\bibliography{mda_papers}

\appendix

\section*{Appendix A: Bias in LDI}
LDI’s false negative rate (FNR) lowers its success when the non-target set is much larger than the target. Specifically, the estimated domain is:
\[
  n_{\text{estimated}} 
  = \text{TPR} \cdot n_{\text{target}} 
    + \text{FNR} \cdot n_{\text{non-target}},
\]
and the attack success rate is:
\[
  \text{Success Rate} 
  = \frac{
      \text{TPR} \cdot n_{\text{target}}
    }{
      \text{TPR} \cdot n_{\text{target}}
      + \text{FNR} \cdot n_{\text{non-target}}
    }.
\]
When \(n_{\text{non-target}} \gg n_{\text{target}}\), even a modest FNR sharply reduces the success rate. An experiment on half of CIFAR-10, MNIST, and EMNIST (the other half plus extra datasets in the pool) confirms this effect, especially given EMNIST’s large size. A reduced EMNIST or LiRA-based ADI partly alleviates the problem.

\begin{figure}[h!]
    \centering
    \begin{subfigure}[b]{0.23\textwidth}
        \centering
        \begin{tikzpicture}
            \begin{axis}[
                ybar,
                bar width=.15cm,
                width=\textwidth,
                height=0.8\textwidth,
                enlargelimits=0.15,
                legend style={at={(0.5,1.05)},
                anchor=south,legend columns=-1},
                symbolic x coords={C-10,M,EM},
                ytick={0.2,0.4,0.6,0.8},
                xtick=data,
                ylabel={Success Rate},
                xlabel={Target domain},
                x tick label style={rotate=45, anchor=east},
                xlabel style={yshift=-10pt},
                every node near coord/.append style={font=\huge, yshift=2pt}
                ]
                \addplot+[error bars/.cd, y dir=both, y explicit]
                    table[x=Datasets,y=Success_rate_HDI,y error=std_HDI, col sep=comma] {results/HDI_bias.csv};
                \addplot+[error bars/.cd, y dir=both, y explicit]
                    table[x=Datasets,y=Success_rate_ADI,y error=std_ADI, col sep=comma] {results/HDI_bias.csv};
                \legend{LDI, LiRA-based-ADI}
            \end{axis}
        \end{tikzpicture}
        \centering
        \caption{Bias data pool}
        \label{exp:bias}
    \end{subfigure}
    \begin{subfigure}[b]{0.23\textwidth}
        \centering
        \begin{tikzpicture}
            \begin{axis}[
                ybar,
                bar width=.15cm,
                width=\textwidth,
                height=0.8\textwidth,
                enlargelimits=0.15,
                legend style={at={(0.5,1.05)},
                anchor=south,legend columns=-1},
                symbolic x coords={C-10,M,EM},
                ytick={0.2,0.4,0.6,0.8},
                xtick=data,
                xlabel={Target domain},
                x tick label style={rotate=45, anchor=east},
                xlabel style={yshift=-10pt},
                every node near coord/.append style={font=\huge, yshift=2pt}
                ]
                \addplot+[error bars/.cd, y dir=both, y explicit]
                    table[x=Datasets,y=Success_rate_HDI,y error=std_HDI, col sep=comma] {results/HDI_balance.csv};
                \addplot+[error bars/.cd, y dir=both, y explicit]
                    table[x=Datasets,y=Success_rate_ADI,y error=std_ADI, col sep=comma] {results/HDI_balance.csv};
            \end{axis}
        \end{tikzpicture}
        \centering
        \caption{Balanced data pool}
        \label{exp:balance}
    \end{subfigure}
    \caption{Success rate of target domains shows that the bias of the data pool significantly impacts the dataset.}
\end{figure}
Experimental results support our theoretical analysis. As shown in Figure \ref{exp:bias}, if we do not reduce the size of EMNIST, the success rates of LDI for CIFAR-10 and MNIST are significantly decreased, while the success rate for EMNIST is much higher. In contrast, ADI demonstrates resistance to the bias in the data pool. Figure \ref{exp:balance} presents the results of both attacks with the size of EMNIST reduced to be similar to that of CIFAR-10 and MNIST. 

\section*{Appendix B: Impact of Shadow Models.}

\begin{table}[h!]
    \centering
    \begin{subtable}[h!]{\linewidth} % Adjust the width as necessary
        \centering
        \begin{tabular}{|c|c|}
            \hline
            \textbf{Concept} & \textbf{Times} \\ \hline
            work\_ware-clothing-covering-artifact-whole & 1 \\ \hline
            skirt-clothing-covering-artifact-whole & 1 \\ \hline
            swimsuit-clothing-covering-artifact-whole & 1 \\ \hline
            garment-clothing-covering-artifact-whole & 7 \\ \hline
        \end{tabular}
        \caption{With concept hierarchy.}
        \label{subtab:a}
    \end{subtable}
    \begin{subtable}[h!]{\linewidth} % Adjust the width as necessary
        \centering
        \begin{tabular}{|c|c|}
            \hline
            \textbf{Concept} & \textbf{Times} \\ \hline
            fur\_coat & 2 \\ \hline
            jeans & 3 \\ \hline
            life\_boat & 2 \\ \hline
            Restaurant & 1 \\ \hline
            Mask & 2 \\ \hline
        \end{tabular}
        \caption{Without concept hierarchy.}
        \label{subtab:b}
    \end{subtable}
    \caption{The quality of identified concepts.}
    \label{tab:concept_times}
\end{table}

\begin{figure*}[h]
   \centering
   \includegraphics[width=1.3\columnwidth]{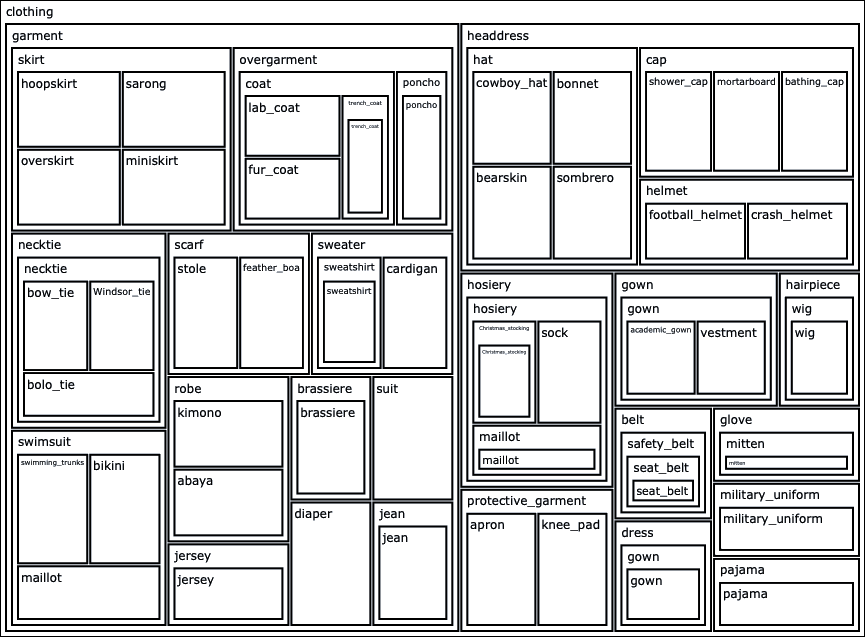}
   \caption{Lower level concept hierarchy of ImageNet-1K.}
   \label{fig:concept2}
\end{figure*}
\begin{figure}[h!]
    \centering
    \begin{tikzpicture}
        \begin{axis}[
            ybar,
            bar width=.15cm,
            width=0.32\textwidth,
            height=0.32\textwidth,
            enlargelimits=0.15,
            legend style={at={(0.5,1.05)}, anchor=south,legend columns=-1},
            symbolic x coords={C-10,M,FM,Pool},
            ytick={0.2,0.4,0.6,0.8},
            xtick=data,
            ylabel={Mix-10 OTDD},
            xlabel={Shadow model},
            x tick label style={rotate=45, anchor=east},
            xlabel style={yshift=-10pt},
            every node near coord/.append style={font=\huge, yshift=2pt}
            ]
            \addplot+[error bars/.cd, y dir=both, y explicit]
                table[x=Datasets,y=OTDD_HDI,y error=std_HDI, col sep=comma] {results/Shadow.csv};
            \addplot+[error bars/.cd, y dir=both, y explicit]
                table[x=Datasets,y=OTDD_ADI,y error=std_ADI, col sep=comma] {results/Shadow.csv};
            \legend{LiRA-ADI, LDI}
        \end{axis}
    \end{tikzpicture}
    \caption{Influence of shadow models on Mix-10 OTDD. "Pool" indicates the shadow model is trained on the entire data pool. Shadow models trained on specific sub-domains decrease the performance of both methods and show higher variability (larger error bars) compared to training shadow models on the entire dataset.}
    \label{exp:shadow}
\end{figure}

In Section \ref{sec:setup}, we introduced membership inference attacks that depend on training multiple shadow models for the target domain. Under domain inference, adversaries do not know the target domain, so we train shadow models on the entire data pool. One might wonder if restricting training to sub-domains (e.g., only CIFAR or only MNIST) would still be effective. As shown in \cite{kaya2021does}, membership inference can remain viable even if shadow models are not trained on the exact target domain. Thus, we hypothesize that ADI and LDI should also provide meaningful results in such scenarios.

To test this, we trained shadow models solely on CIFAR-10, MNIST, and Fashion-MNIST, while our target domain was a “Mix-10” set of random classes from the synthesized datasets. Figure \ref{exp:shadow} demonstrates that training on the full data pool is more effective. Limiting training to one sub-domain lowers attack performance and increases its uncertainty due to the randomness of target domains. For instance, if Mix-10 consists mainly of CIFAR-like classes, shadow models trained on MNIST-like domains perform poorly.

\section*{Appendix C: Straightforward Results on ImageNet Concept Hierarchy}

We present the straightforward results of LiRA-based ADI attacks on the DeepFashion-trained model. The attack was repeated 10 times, identifying the branch with the highest probability as the model's estimated domain. Results with the concept hierarchy (Table \ref{subtab:a}) show the attacker accurately deduces the domain as clothing-related items. Without the hierarchy (Table \ref{subtab:b}), the estimation is less precise, with non-clothing items like 'restaurant' and 'mask' causing ambiguity. To further clarify, Figures \ref{fig:concept2} visualize the concept hierarchies used in our experiments.

\section*{Appendix D: Results on Synthesized Datasets}
The results in Figures \ref{fig:OTDD_mix_barplot}, \ref{fig:MI_mix_barplot}, and \ref{fig:access_mix_barplot} show that LiRA-based ADI performs the best, followed by entropy-based ADI and LDI. GDI produces meaningful results only on single datasets.

\begin{figure*}[h!]
    \centering
    \begin{subfigure}[b]{0.32\textwidth}
        \centering
        \begin{tikzpicture}
            \begin{axis}[
                ybar,
                bar width=.1cm,
                width=0.8\textwidth,
                height=0.8\textwidth,
                enlargelimits=0.15,
                legend style={at={(0.5,1.2)}, anchor=north, legend columns=-1, yshift=-10pt},
                ylabel={OTDD},
                symbolic x coords={Mix-10, Mix-20, Mix-30},
                xtick=data,
                xlabel={Target domain},
                x tick label style={rotate=45, anchor=east},
                xlabel style={yshift=-15pt},
                every node near coord/.append style={font=\huge, yshift=2pt}
                ]
                \addplot+[error bars/.cd, y dir=both, y explicit]
                    table[x=Datasets, y=ADI, y error=ADI_std, col sep=comma] {results/OTDD_mix.csv};
                \addplot+[error bars/.cd, y dir=both, y explicit]
                    table[x=Datasets, y=ADI_E, y error=ADI_std_E, col sep=comma] {results/OTDD_mix.csv};
                \addplot+[error bars/.cd, y dir=both, y explicit]
                    table[x=Datasets, y=HDI, y error=HDI_std, col sep=comma] {results/OTDD_mix.csv};
                \addplot+[error bars/.cd, y dir=both, y explicit]
                    table[x=Datasets, y=GDI, y error=GDI_std, col sep=comma] {results/OTDD_mix.csv};
            \end{axis}
        \end{tikzpicture}
         
        \caption{Synthesized datasets}
        \label{exp:mix_otdd}
    \end{subfigure}
    \hfill
    \begin{subfigure}[b]{0.32\textwidth}
        \centering
        \begin{tikzpicture}
            \begin{axis}[
                ybar,
                bar width=.06cm,
                width=1.4\textwidth,
                height=0.8\textwidth,
                enlargelimits=0.15,
                legend style={at={(0.5,1.2)}, anchor=north, legend columns=-1, yshift=-10pt},
                symbolic x coords={M, E, L, C-10, C-100, CN-10, FM, CL},
                xtick=data,
                xlabel={Target domain},
                x tick label style={rotate=45, anchor=east},
                xlabel style={yshift=-15pt},
                every node near coord/.append style={font=\huge, yshift=2pt}
                ]
                \addplot+[error bars/.cd, y dir=both, y explicit]
                    table[x=Datasets, y=ADI, y error=ADI_std, col sep=comma] {results/OTDD_single.csv};
                \addplot+[error bars/.cd, y dir=both, y explicit]
                    table[x=Datasets, y=ADI_E, y error=ADI_std_E, col sep=comma] {results/OTDD_single.csv};
                \addplot+[error bars/.cd, y dir=both, y explicit]
                    table[x=Datasets, y=HDI, y error=HDI_std, col sep=comma] {results/OTDD_single.csv};
                \addplot+[error bars/.cd, y dir=both, y explicit]
                    table[x=Datasets, y=GDI, y error=GDI_std, col sep=comma] {results/OTDD_single.csv};
                \legend{LiRA\_ADI, Entropy\_ADI, HDI, GDI}
            \end{axis}
        \end{tikzpicture}
         
        \caption{Single datasets}
        \label{exp:single_otdd}
    \end{subfigure}
    \hfill
    \begin{subfigure}[b]{0.32\textwidth}
        \centering
        \begin{tikzpicture}
            \begin{axis}[
                ybar,
                bar width=.1cm,
                width=0.8\textwidth,
                height=0.8\textwidth,
                enlargelimits=0.15,
                legend style={at={(0.5,1.2)}, anchor=north, legend columns=-1, yshift=-10pt},
                symbolic x coords={NETTE, Woof, Fashion},
                xtick=data,
                xlabel={Target domain},
                x tick label style={rotate=45, anchor=east},
                xlabel style={yshift=-15pt},
                every node near coord/.append style={font=\huge, yshift=2pt}
                ]
                \addplot+[error bars/.cd, y dir=both, y explicit]
                    table[x=Datasets, y=ADI, y error=ADI_std, col sep=comma] {results/OTDD_high.csv};
                \addplot+[error bars/.cd, y dir=both, y explicit]
                    table[x=Datasets, y=ADI_E, y error=ADI_std_E, col sep=comma] {results/OTDD_high.csv};
                \addplot+[error bars/.cd, y dir=both, y explicit]
                    table[x=Datasets, y=HDI, y error=HDI_std, col sep=comma] {results/OTDD_high.csv};
                \addplot+[error bars/.cd, y dir=both, y explicit]
                    table[x=Datasets, y=GDI, y error=GDI_std, col sep=comma] {results/OTDD_high.csv};
            \end{axis}
        \end{tikzpicture}
         
        \caption{ImageNet-related datasets}
        \label{exp:high_otdd}
    \end{subfigure}
     
    \caption{OTDD between the extracted dataset and the target domain.}
    \label{fig:OTDD_mix_barplot}
\end{figure*}

\begin{figure*}[h!]
    \centering
    \begin{subfigure}[b]{0.32\textwidth}
        \centering
        \begin{tikzpicture}
            \begin{axis}[
                ybar,
                bar width=.1cm,
                width=0.8\textwidth,
                height=0.8\textwidth,
                enlargelimits=0.15,
                legend style={at={(0.5,1.2)},
                anchor=north,legend columns=-1, yshift=-10pt},
                ylabel={MI Acc},
                symbolic x coords={Mix-10, Mix-20, Mix-30},
                xtick=data,
                xlabel={Target domain},
                x tick label style={rotate=45, anchor=east},
                xlabel style={yshift=-15pt},
                every node near coord/.append style={font=\huge, yshift=2pt}
                ]
                \addplot+[error bars/.cd, y dir=both, y explicit]
                    table[x=Datasets,y=Baseline,y error=std_baseline, col sep=comma] {results/MI_mix.csv};
                \addplot+[error bars/.cd, y dir=both, y explicit]
                    table[x=Datasets,y=ADI,y error=std_ADI, col sep=comma] {results/MI_mix.csv};
                     \addplot+[error bars/.cd, y dir=both, y explicit]
                    table[x=Datasets,y=ADI_E,y error=std_ADI_E, col sep=comma] {results/MI_mix.csv};
                \addplot+[error bars/.cd, y dir=both, y explicit]
                    table[x=Datasets,y=HDI,y error=std_HDI, col sep=comma] {results/MI_mix.csv};
                \addplot+[error bars/.cd, y dir=both, y explicit]
                    table[x=Datasets,y=GDI,y error=std_GDI, col sep=comma] {results/MI_mix.csv};
            \end{axis}
        \end{tikzpicture}        
                                                       
        \caption{Synthesized datasets}
        \label{exp:mix_MI}
    \end{subfigure}
    \hfill
    \centering
    \begin{subfigure}[b]{0.32\textwidth}
        \centering
        \begin{tikzpicture}
            \begin{axis}[
                ybar,
                bar width=.05cm,
                width=1.4\textwidth,
                height=0.8\textwidth,
                enlargelimits=0.15,
                legend style={at={(0.5,1.4)},
                anchor=north,legend columns=3, yshift=-10pt},
                symbolic x coords={M,E,L,C-10,C-100,CN-10,FM,CL},
                xtick=data,
                xlabel={Target domain},
                x tick label style={rotate=45, anchor=east},
                xlabel style={yshift=-15pt},
                every node near coord/.append style={font=\huge, yshift=2pt}
                ]
                \addplot+[error bars/.cd, y dir=both, y explicit]
                    table[x=Datasets,y=Baseline,y error=Baseline_std, col sep=comma] {results/MI_single.csv};
                    \addplot+[error bars/.cd, y dir=both, y explicit]
                    table[x=Datasets,y=ADI_E,y error=std_ADI_E, col sep=comma] {results/MI_single.csv};
                \addplot+[error bars/.cd, y dir=both, y explicit]
                    table[x=Datasets,y=ADI,y error=ADI_std, col sep=comma] {results/MI_single.csv};
                \addplot+[error bars/.cd, y dir=both, y explicit]
                    table[x=Datasets,y=HDI,y error=HDI_std, col sep=comma] {results/MI_single.csv};
                \addplot+[error bars/.cd, y dir=both, y explicit]
                    table[x=Datasets,y=GDI,y error=GDI_std, col sep=comma] {results/MI_single.csv};
                \legend{Baseline,LiRA\_ADI, Entropy\_ADI, HDI, GDI}
            \end{axis}
        \end{tikzpicture}
         
        \caption{Single datasets}
        \label{exp:single_MI}
    \end{subfigure}
    \hfill
    \begin{subfigure}[b]{0.32\textwidth}
        \centering
        \begin{tikzpicture}
            \begin{axis}[
                ybar,
                bar width=.1cm,
                width=0.8\textwidth,
                height=0.8\textwidth,
                enlargelimits=0.15,
                legend style={at={(0.5,1.2)},
                anchor=north,legend columns=-1, yshift=-10pt},
                symbolic x coords={NETTE,Woof,Fashion},
                xtick=data,
                xlabel={Target domain},
                x tick label style={rotate=45, anchor=east},
                xlabel style={yshift=-15pt},
                every node near coord/.append style={font=\huge, yshift=2pt}
                ]
                \addplot+[error bars/.cd, y dir=both, y explicit]
                    table[x=Datasets,y=Baseline,y error=std_baseline, col sep=comma] {results/MI_high.csv};
                \addplot+[error bars/.cd, y dir=both, y explicit]
                    table[x=Datasets,y=ADI,y error=std_HDI, col sep=comma] {results/MI_high.csv};
                    \addplot+[error bars/.cd, y dir=both, y explicit]
                    table[x=Datasets,y=ADI_E,y error=std_ADI_E, col sep=comma] {results/MI_high.csv};
                \addplot+[error bars/.cd, y dir=both, y explicit]
                    table[x=Datasets,y=HDI,y error=std_HDI, col sep=comma] {results/MI_high.csv};
                \addplot+[error bars/.cd, y dir=both, y explicit]
                    table[x=Datasets,y=GDI,y error=std_GDI, col sep=comma] {results/MI_high.csv};
            \end{axis}
        \end{tikzpicture}
         
        \caption{ImageNet-related datasets}
        \label{exp:high_MI}
    \end{subfigure}
     
    \caption{Enhancement to model inversion attacks by estimated domain. ADI and HDI perform well in three scenarios. GDI only performs good in the second scenario.}
    \label{fig:MI_mix_barplot}
\end{figure*}

\begin{figure*}[h!]
    \centering
    \begin{subfigure}[b]{0.32\textwidth}
        \centering
        \begin{tikzpicture}
            \begin{axis}[
                ybar,
                bar width=.1cm,
                width=0.8\textwidth,
                height=0.8\textwidth,
                enlargelimits=0.15,
                legend style={at={(0.5,1.2)},
                anchor=north,legend columns=-1, yshift=-10pt},
                ylabel={Accesses},
                xlabel={Target domain},
                symbolic x coords={Mix-10, Mix-20, Mix-30},
                xtick=data,
                x tick label style={rotate=45, anchor=east},
                xlabel style={yshift=-15pt},
                every node near coord/.append style={font=\huge, yshift=2pt}
                ]
                \addplot+[error bars/.cd, y dir=both, y explicit]
                    table[x=Datasets,y=ADI,y error=ADI_std, col sep=comma] {results/access_mix.csv};
                    \addplot+[error bars/.cd, y dir=both, y explicit]
                    table[x=Datasets,y=ADI_E,y error=ADI_std_E, col sep=comma] {results/access_mix.csv};
                \addplot+[error bars/.cd, y dir=both, y explicit]
                    table[x=Datasets,y=HDI,y error=HDI_std, col sep=comma] {results/access_mix.csv};
                \addplot+[error bars/.cd, y dir=both, y explicit]
                    table[x=Datasets,y=GDI,y error=GDI_std, col sep=comma] {results/access_mix.csv};
            \end{axis}
        \end{tikzpicture}        
                                                       
        \caption{Synthesized datasets}
        \label{exp:mix_MI}
    \end{subfigure}
    \hfill
    \begin{subfigure}[b]{0.32\textwidth}
        \centering
        \begin{tikzpicture}
            \begin{axis}[
                ybar,
                bar width=.05cm,
                width=1.4\textwidth,
                height=0.8\textwidth,
                enlargelimits=0.15,
                legend style={at={(0.5,1.2)},
                anchor=north,legend columns=-1, yshift=-10pt},
                symbolic x coords={M,E,L,C-10,C-100,CN-10,FM,CL},
                xtick=data,
                xlabel={Target domain},
                x tick label style={rotate=45, anchor=east},
                xlabel style={yshift=-15pt},
                every node near coord/.append style={font=\huge, yshift=2pt}
                ]
                \addplot+[error bars/.cd, y dir=both, y explicit]
                    table[x=Datasets,y=ADI,y error=ADI_std, col sep=comma] {results/access_single.csv};
                     \addplot+[error bars/.cd, y dir=both, y explicit]
                    table[x=Datasets,y=ADI_E,y error=ADI_std_E, col sep=comma] {results/access_single.csv};
                \addplot+[error bars/.cd, y dir=both, y explicit]
                    table[x=Datasets,y=HDI, col sep=comma] {results/access_single.csv};
                \addplot+[error bars/.cd, y dir=both, y explicit]
                    table[x=Datasets,y=GDI,y error=GDI_std, col sep=comma] {results/access_single.csv};
                \legend{ LiRA\_ADI, Entropy\_ADI, HDI, GDI}
            \end{axis}
        \end{tikzpicture}
         
        \caption{Single datasets}
        \label{exp:single_MI}
    \end{subfigure}
    \hfill
    \begin{subfigure}[b]{0.32\textwidth}
        \centering
        \begin{tikzpicture}
            \begin{axis}[
                ybar,
                bar width=.1cm,
                width=0.8\textwidth,
                height=0.8\textwidth,
                enlargelimits=0.15,
                legend style={at={(0.5,1.2)},
                anchor=north,legend columns=-1, yshift=-10pt},
                symbolic x coords={NETTE,Woof,Fashion},
                xtick=data,
                xlabel={Target domain},
                x tick label style={rotate=45, anchor=east},
                xlabel style={yshift=-15pt},
                every node near coord/.append style={font=\huge, yshift=2pt}
                ]
                \addplot+[error bars/.cd, y dir=both, y explicit]
                    table[x=Datasets,y=ADI,y error=ADI_std, col sep=comma] {results/access_high.csv};
                     \addplot+[error bars/.cd, y dir=both, y explicit]
                    table[x=Datasets,y=ADI_E,y error=ADI_std_E, col sep=comma] {results/access_high.csv};
                \addplot+[error bars/.cd, y dir=both, y explicit]
                    table[x=Datasets,y=HDI, col sep=comma] {results/access_high.csv};
                \addplot+[error bars/.cd, y dir=both, y explicit]
                    table[x=Datasets,y=GDI,y error=GDI_std, col sep=comma] {results/access_high.csv};
            \end{axis}
        \end{tikzpicture}
         
        \caption{ImageNet-related datasets}
        \label{exp:high_MI}
    \end{subfigure}
     
    \caption{Number of access. ADI needs the smallest amount of access. Both HDI and GDI suffer from huge amounts of access.}
    \label{fig:access_mix_barplot}
\end{figure*}

\end{document}